\documentclass{ecai} 

\usepackage{latexsym}
\usepackage{amssymb}
\usepackage{amsmath}
\usepackage{amsthm}
\usepackage{booktabs}
\usepackage{enumitem}
\usepackage{graphicx}
\usepackage{color}

\usepackage{svg}

\newcommand{\BibTeX}{B\kern-.05em{\sc i\kern-.025em b}\kern-.08em\TeX}

\begin{document}


\begin{frontmatter}


\paperid{3306} 


\title{GRASPED: Graph Anomaly Detection using Autoencoder with Spectral Encoder and Decoder}


\author[A]{\fnms{Wei Herng}~\snm{Choong}\footnote{Equal contribution.}}
\author[A, B]{\fnms{Jixing}~\snm{Liu}\thanks{Corresponding Author. Email: jixing.liu@tum.de}\footnotemark[1]}
\author[A]{\fnms{Ching-Yu}~\snm{Kao}}
\author[A]{\fnms{Philip}~\snm{Sperl}} 

\address[A]{Fraunhofer Institute for Applied and Integrated Security (AISEC), Garching near Munich, Germany}
\address[B]{Technical University of Munich, Munich, Germany}


\begin{abstract}
Graph machine learning has been widely explored in various domains, such as community detection, transaction analysis, and recommendation systems. In these applications, anomaly detection plays an important role. Recently, studies have shown that anomalies on graphs induce spectral shifts. Some supervised methods have improved the utilization of such spectral domain information. However, they remain limited by the scarcity of labeled data due to the nature of anomalies. On the other hand, existing unsupervised learning approaches predominantly rely on spatial information or only employ low-pass filters, thereby losing the capacity for multi-band analysis. In this paper, we propose Graph Autoencoder with Spectral Encoder and Spectral Decoder (GRASPED) for node anomaly detection. Our unsupervised learning model features an encoder based on Graph Wavelet Convolution, along with structural and attribute decoders. The Graph Wavelet Convolution-based encoder, combined with a Wiener Graph Deconvolution-based decoder, exhibits bandpass filter characteristics that capture global and local graph information at multiple scales. This design allows for a learning-based reconstruction of node attributes, effectively capturing anomaly information. Extensive experiments on several real-world graph anomaly detection datasets demonstrate that GRASPED outperforms current state-of-the-art models.
\end{abstract}

\end{frontmatter}


\section{Introduction}

Graph anomaly detection (GAD) refers to the process of identifying unusual instances, whether nodes or edges, within a graph that exhibit properties deviating significantly from the given notion of normal. Given that graph structures can be frequently found in real-world scenarios, GAD has a wide range of applications across various domains, including financial systems \cite{zhang2021fraudre,zhang2022efraudcom}, social networks \cite{Heard_2010,savage2016anomalydetectiononlinesocial}, or network security \cite{zhong2024survey}. 

Anomaly detection on graphs presents some unique challenges. For example, graphs contain not only attribute information from the nodes and edges but also topological information that describes the relationship among nodes. Furthermore, there are diverse types of graphs. Graphs can be classified as static or dynamic. They may exhibit heterophilic or homophilic characteristics, depending on the nature of the relationship between nodes. Hence, formulating a unified definition for graph anomalies is complex. For anomaly detection in static attributed graphs, \cite{liu2022bond} identified two groups of anomalies: structural and contextual. They defined structural anomalies as densely connected nodes and contextual anomalies as nodes that exhibit significantly different attributes compared to their neighboring nodes. However, these definitions often do not apply in real-world datasets, as anomalous samples could also be a mixture of structural and contextual components. Hence, it is important that the anomaly-detection model captures as much information as possible from the graph during detection. 

Graph neural networks (GNNs) are designed to extract complex graph characteristics. Their ability to learn meaningful representations using structural and attribute information from graph nodes leads them to outperform traditional anomaly detection approaches \cite{liu2022bond}. GNNs can be grouped into two categories: spatial and spectral GNNs. Spatial GNNs rely on the message-passing mechanism to learn graph representations. The representation is learned by aggregating information from the neighboring nodes. As the message-passing mechanism is based on the homogeneity assumption of a graph, they exhibit limitations when they are applied to graphs that do not fulfill this assumption, e.g., a heterophilic graph. Existing methods that aim to elevate this problem include applying an attention mechanism \cite{liu2021intention,wang2019semi} or performing neighborhood resampling \cite{liu2021pick,liu2020alleviating}. On the other hand, spectral GNNs tackle this problem by leveraging specially designed filters that process the Laplacian matrix of graphs. Anomaly detection on spectral graphs requires spectral filters to provide a high learning capability. Previous work \cite{tang2022rethinking} suggests that anomalies cause a ``right-shift'' phenomenon in the spectral energy distribution, where the energy shifts to the higher frequency regions. Hence, graph spectral filters that could not capture high-frequency components of the spectral graph, such as filters with low polynomial bases, would fail to detect the anomalies that induce the ``right-shift'' phenomenon.  In our work, we leverage the high learning capability of filters \cite{xu2019graph} to capture high-frequency signals on spectral graphs.

Autoencoders are widely used for anomaly detection \cite{zhou2017anomaly,sakurada2014anomaly,chen2018autoencoder}. They aim to learn the compressed representation of the input data, and reconstruct the data based on the compressed representation with minimal reconstruction loss. In anomaly detection, the reconstruction loss is used as a metric to determine whether an input is an anomaly. An input is considered anomalous when the reconstruction loss, or the anomaly score, is higher than a predefined threshold. In GAD, Graph Autoencoders (GAEs) leverage GNNs to extract attribute and structural information from graphs to detect anomalies. For node anomaly detection, each of the reconstructed nodes is assigned an anomaly score. Nodes with a score higher than a predefined threshold are considered anomalies. Most current GAEs for node anomaly detection use spatial GNNs as encoders~\cite{kipf2016semi,kipf2016variational}. Furthermore, they rely on Multi-Layer Perceptron (MLP)-based~\cite{roy2024gad} and/or inner product decoders for graph reconstruction, which have lower expressive power. In our work, we explore the usage of a spectral-based GNN encoder and incorporate a graph deconvolutional network as our decoder. Our experiments on five different real-world datasets show that GRASPED outperforms current state-of-the-art models in node anomaly detection.

In summary, we propose our anomaly detection approach called \textbf{GRASPED}, which consists of a \textbf{GRA}ph Autoencoder with a \textbf{SP}ectral \textbf{E}ncoder and \textbf{D}ecoder. Our approach is a GAE-based anomaly detection model that captures multi-resolution spectral information and leverages graph deconvolution as a powerful decoder to complement GNN-based encoders.
Inspired by advances in temporal signal and image data processing, GRASPED adopts a paradigm that first decomposes graph data using graph wavelet transforms and subsequently reconstructs it from filtered latent embeddings. Specifically, GRASPED encodes vertex-domain information into spectral-domain latent embeddings via a GWNN \cite{xu2019graph}, enabling multi-resolution spectral capture. These embeddings are then reconstructed into the vertex domain through two decoders: a Graph Deconvolution Neural Network (GDN) \cite{li2021deconvolutional} for attribute reconstruction and an MLP-based structural decoder for adjacency matrix reconstruction. The GWNN acts as a learnable band-pass filter, capturing both low- and high-frequency spectral information at multiple scales. Meanwhile, the GDN decoder ensures expressive reconstruction of features, addressing the limitations of traditional inner-product or MLP-based decoders. This complementary design bridges the gap by leveraging information from the spectral domain for effective graph anomaly detection.


\section{Related Work}

Graph Convolutional Networks (GCNs)~\cite{kipf2016semi} represent one of the earliest instances of spectral domain GNNs, utilizing the graph Laplacian matrix to process graph signals. However, due to their reliance on low-order polynomial approximations, GCNs inherently behave as low-pass filters, limiting their ability to capture high-frequency components in the spectral domain~\cite{nt2019revisiting}. To address this, Xu et al.~\cite{xu2019graph} introduced the Graph Wavelet Neural Network (GWNN) based on Hammond’s graph wavelet theory~\cite{hammond2011wavelets}, employing polynomial-based methods to capture multi-resolution spectral information more effectively. In addition, Yang et al.~\cite{yang2024wavenet} leveraged Multi-Resolution Analysis (MRA) to better model graph signals, proposing WaveNet — a  GNN designed to more accurately capture the high-frequency components. Tang et al.~\cite{tang2022rethinking} observed that anomalies induce a “right-shift” phenomenon in the spectral energy distribution, where energy becomes more concentrated in higher frequencies and diminished in lower frequencies. Since  GNNs act as localized band-pass filters~\cite{nt2019revisiting}, they are naturally better suited to handle these spectral changes compared to conventional GNNs such as GCN~\cite{kipf2016semi} and ChebyNet~\cite{defferrard2016convolutional}, which primarily function as low-pass filters [16]. To address this challenge, Tang et al. proposed the Beta Wavelet Graph Neural Network (BWGNN), a novel variant of GWNN specifically tailored for detecting anomalies by effectively capturing the altered spectral patterns~\cite{tang2022rethinking}.

Graph autoencoders (GAEs), introduced by Kipf and Welling~\cite{kipf2016variational}, combine GNNs with autoencoder architectures. These models encode graph data and aim to reconstruct it, leveraging reconstruction errors as anomaly scores. Nodes with higher reconstruction errors are considered more likely to be anomalous. AnomalyDAE, proposed by Fan et al.~\cite{fan2020anomalydae}, captures cross-modality interactions between graph structures and node attributes. Using Graph Attention Networks~(GAT)~\cite{velivckovic2017graph} and MLP as encoders, it reconstructs both the adjacency matrix and the attribute matrix via structure and attribute decoders. This dual-autoencoder design enhances anomaly detection from both structural and contextual perspectives. However, its inner product-based decoders are not learnable, limiting their expressive power. GAD-NR employs GCN-like models as encoders and MLP-based decoders to capture structural and contextual information~\cite{roy2024gad}. Its learning-based decoders outperform inner product-based approaches in expressiveness, which contributed to its recognition as a prior state-of-the-art (SOTA) model. However, MLP-based decoders fail to fully exploit graph spectral information, leaving room for improvement in leveraging graph-rich structural and feature information. ADA-GAD (Anomaly-Denoised Autoencoders for Graph Anomaly Detection), proposed by He et al.~\cite{he2024ada}, is a two-stage GAE framework inspired by the ``right-shift'' phenomenon. Its first stage involves anomaly-denoised augmentation, followed by retraining in the second stage using multi-level representations. However, similar to GAD-NR, it employs MLP-based decoders, which limit its ability to symmetrically handle spectral information during reconstruction.


\section{Preliminaries}

We consider the scenario of node-level anomaly detection on an undirected graph $\mathcal{G}=(\mathcal{V}, \mathcal{E}, \mathbf{X})$ with node set $\mathcal{V}$ and edge set $\mathcal{E}$. In spectral GNNs, the feature matrix $\mathbf{X} \in \mathbb{R}^{n \times d}$ is treated as graph signals, where $n = |\mathcal{V}|$ and $\boldsymbol{x}(i)$ denotes the signal of dimension $d$ from node $i$. 
The structural information of a graph is denoted as an adjacency matrix $\mathbf{A}$, and $\mathbf{D}$ denotes the diagonal degree matrix. The normalized adjacency matrix can be represented as $\tilde{\mathbf{A}}=$ $\mathbf{D}^{-1 / 2} \mathbf{A} \mathbf{D}^{-1 / 2}$. With that, we can derive the normalized Laplacian matrix as $\mathbf{L}=\mathbf{I}-\tilde{\mathbf{A}}$, with $\mathbf{I}$ representing the identity matrix. The eigenvalues $\lambda_{i}$ of $\mathbf{L}$ lie in the interval $\lambda_{i} \in[0,2]$. Let $\mathbf{L}=\mathbf{U} \boldsymbol{\Lambda} \mathbf{U}^{\top}$ denote the eigendecomposition of a normalized Laplacian matrix, where $\mathbf{U}$ is the eigenvector matrix, $\boldsymbol{\Lambda}=\operatorname{diag}\left[\lambda_{1}, \ldots, \lambda_{n}\right]$ is the diagonal eigenvalue matrix.

\subsection{Graph Convolution via Signal Filter and Spectral GNNs}
The graph Fourier transform of $\mathbf{X}$ is defined as $\hat{\mathbf{X}}=\mathbf{U}^{\top} \mathbf{X}$, and the inverse transform is $\mathbf{X}=\mathbf{U} \hat{\mathbf{X}}$ \cite{shuman2013emerging}. When filtering on spectral graph signals, we utilize a filter function $g_{c}(\cdot)$ on the Laplacian matrix. This can be described as $g_{c}(\mathbf{L}) \mathbf{X} = \mathbf{U} g_{c}(\boldsymbol{\Lambda}) \mathbf{U}^{\top} \mathbf{X}$.  Details of $g_{c}(\cdot)$ is explained in Section \ref{sec:ge}.

\subsection{Graph Deconvolution Using Inverse Function of Signal Filter}
Graph deconvolution operates as the inverse of graph convolution, designed to recover original node attributes from their smoothed representations. From a spectral perspective, this process reverses the convolution filter's effect in the graph Fourier domain \cite{li2021deconvolutional}. Formally, given a smoothed signal \(\mathbf{h} \in \mathbb{R}^N\) and a deconvolution filter \(g_d\), the operation is defined as
$\hat{\mathbf{x}} = \mathbf{U} \, g_d(\mathbf{\Lambda}) \, \mathbf{U}^\top \mathbf{h}$. Equivalently, this can be written as $\hat{\mathbf{x}} = g_d(\mathbf{L}) \mathbf{h}$ using the Laplacian operator $\bf{L}$. The deconvolution filter \(g_d\) is typically chosen as the inverse of the original convolution filter \(g_c\). In this paper we use \textbf{Heat kernel diffusion}: The deconvolution filter \(g_d(\lambda_i) = e^{\lambda_i}\) counteracts the smoothing from \(g_c(\lambda_i) = e^{-\lambda_i}\).

\section{Methodology}

\begin{figure}[htbp]
  \centering
  \includegraphics[width=0.95\columnwidth]{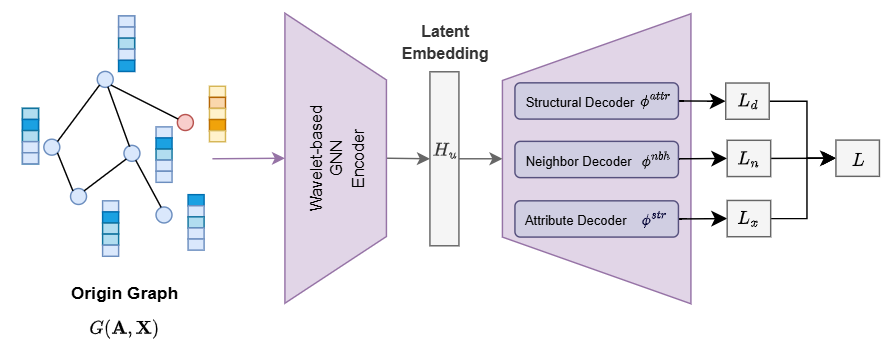}
  \caption{Illustration of GRASPED}
  \label{fig:GRASPED_overview}
\end{figure}
In this section, we present the details of our graph anomaly detection method called GRASPED.
We divide the presentation into three parts: First, we show the encoding part which consists of a graph encoder. 
Second, we describe the decoding part of our system.
Finally, we elaborate on the optimization objective of our anomaly detection approach.
An overview of our method is shown in Figure \ref{fig:GRASPED_overview}.

\subsection{Graph Encoder}
\label{sec:ge}
This section presents a multiscale spectral graph neural network framework that integrates wavelet-based MRA with adaptive filter learning. By hierarchically combining scaling functions and leveraging refinement relations, our approach constructs interpretable graph filters. We focus on the graph signal space \(L^2(\mathcal{G})\)\footnote{The space $L^2(G)$, also known as the square-integrable function space, consists of all functions $f(t)$ defined on the real line such that the integral of their squared magnitude is finite. Such functions are called finite-energy functions. $L^2(G)$ is used to handle finite-energy signals and serves as the foundational space for wavelet analysis and MRA.}, decomposed into nested subspaces through MRA. The signal space decomposes into a sequence of approximation spaces \(\{V_j\}_{j \in \mathbb{Z}}\) and detail spaces \(\{W_j\}_{j \in \mathbb{Z}}\). \(V_j\) captures low-frequency components at resolution \(j\), and \(W_j\) encodes high-frequency details. Then we have a father scaling function $\phi_{j,k}$ and a mother wavelet function $\psi_{j,k}$. The father scaling function $\phi_{j,k}$ spans \(V_j\) and satisfies the \textit{refinement relation}:

    \begin{equation}
        \phi_{j,k}(x) = \sum_{m \in \mathbb{Z}} h_{j,m} \phi_{j+1,2k+m}(x),
        \label{eq:father_refinement}
    \end{equation}
where \(h_{j,m}\) are refinement coefficients. The mother wavelet function $\psi_{j,k}$ spans \(W_j\) and is derived from scaling functions:
    \begin{equation}
        \psi_{j,k}(x) = \sum_{m \in \mathbb{Z}} g_{j,m} \phi_{j+1,2k+m}(x),
        \label{eq:mother_refinement}
    \end{equation}
where \(g_{j,m}\) are wavelet coefficients. Based on the Mallat algorithm \cite{mallat1999wavelet}, any signal \( f \in L^2(\mathcal{G}) \) admits:
\begin{align}
    f = \underbrace{\sum_k \alpha_{0,k} \phi_{0,k}}_{\text{Coarse approximation}} + \sum_{j=1}^\infty \underbrace{\sum_k \beta_{j,k} \psi_{j,k}}_{\text{Detail components}}\,,
    \label{eq:wavelet_decomp}
\end{align}
where \( \alpha_{j,k} = \langle f, \phi_{j,k} \rangle \), \( \beta_{j,k} = \langle f, \psi_{j,k} \rangle \). As we discussed above, lower-resolution bases can be constructed from higher-resolution scaling functions through linear combinations. Combining Equation \ref{eq:mother_refinement} and Equation \ref{eq:wavelet_decomp}, we obtain:
\begin{align}
f &= \sum_k \alpha_{0,k}\,\phi_{0,k}
     + \sum_{j=0}^\infty \sum_k \beta_{j,k} 
     \left(\sum_{m \in \mathbb{Z}} g_{j,m}\,\phi_{j+1,2k+m}\right) \nonumber \\
  &= \sum_k \alpha_{0,k}\,\phi_{0,k}
     + \sum_{j=0}^{\infty} \sum_{2k + m} \left\langle f, \phi_{j+1, 2k+m} \right\rangle \phi_{j+1, 2k+m} \nonumber\\
  &= \sum_k \alpha_{0,k}\,\phi_{0,k}
     + \sum_{j=1}^\infty \sum_{2k+m} \gamma_{j,2k+m}\,\phi_{j,2k+m}\,,
     \label{eq:wavelet_decompe_derive1}
\end{align}
where $\gamma_{\cdot,\cdot}$ represents the inner product coefficient obtained by projecting the function $f$ onto the finer-scale basis functions $\phi_{\cdot,\cdot} $ and $m\in\{0, 1, 2, \dots, L_{\text{f}} - 1\}$ corresponds to the shift of the filter coefficients, with \( L_{\text{f}} \) representing the filter length. In practice, for real‐world (finite) signals or when only a finite approximation accuracy is required, one truncates the infinite wavelet decomposition at \(j = J - 1\). This corresponds to performing \(J\) decomposition levels in total. Consequently, at the finest (or highest‐frequency) level \(j = J - 1\), the total number of shift positions is \(K = 2^{J-1}\).

Combining Equation \ref{eq:father_refinement} and Equation \ref{eq:wavelet_decompe_derive1}, the spectral filter \(g_c(\lambda)\) is parametrized as a linear combination of Haar scaling functions across \(J\) decomposition layers:
\begin{equation}
    g_c(\lambda) = \sum_{j=0}^{J-1} \sum_{k=0}^{2^j-1} \gamma_{j,k}\,\phi_{\text{Haar},j,k}(\lambda)
     = \sum_{k=0}^{K-1} \theta_{J,k}\,\phi_{\text{Haar},J,k}(\lambda),
    \label{eq:spectral_filter}
\end{equation}
where \(\theta_{J,k} \in \mathbb{R}\) are learnable coefficients. We adopt Haar bases, whose filter length \(L_{f}\) is 2,  for their simplicity, orthogonality, and computational efficiency. We substitute the spectral filter \(g_c(\lambda)\) into spectral convolution and get:
\begin{equation}
    \mathbf{H} = \mathbf{U} \mathbf{G}_c\mathbf{U}^\top \mathbf{X}\mathbf{W},
    \label{eq:spectral_convolution}
\end{equation}
with \(\mathbf{G}_c = \text{diag}\left(g_c(\lambda_1), \ldots, g_c(\lambda_n)\right)\). We can encode the spatial realization of \(g_c(\lambda)\) in the Multiscale Diffusion Operator \(\mathbf{M}\) by:
\begin{equation}
    \mathbf{M} = \mathbf{U} \mathbf{G}_c\mathbf{U}^\top.
    \label{eq:diffusion_operator}
\end{equation}
Finally, the output of the \((i)\)-th layer of \(Z\)-layer wavelet-based Graph Encoder \(\mathbf{H_u^{(i)}}\) is equivalently expressed as:
\begin{equation}
    \mathbf{H_u^{(i)}} = \sigma \left(\mathbf{M} \mathbf{H_u^{(i-1)}}\mathbf{W^{(i-1)}} \right),
    \label{eq:spatial_convolution}
\end{equation}
where \(\mathbf{H_u^{(0)}} = \mathbf{X}\), and \(\sigma\) is the activation function such as ReLU. 

\subsection{Structural, Neighbor, and Attribute Decoders}
As shown in Figure \ref{fig:GRASPED_overview}, our decoder consists of three components. We follow the idea of neighborhood-based reconstruction from GAD-NR~\cite{roy2024gad}: a structure decoder \(\phi_{str}(\cdot)\) and a neighbor decoder \(\phi_{nbh}(\cdot)\), both utilizing the latent embedding \(\mathbf{h}_u^{(Z)}\) from the final layer of the wavelet-based GNN Encoder with \(Z\) layers. We use a powerful GDN for attribute reconstruction. \(\mathbf{h}_u^{(Z)}\) serves as the latent representation of node \(u\) in the following sections.  

\subsubsection{Structure Decoder}  
For node degree reconstruction, we design an MLP \(\phi_{str}(\cdot)\) that maps \(\mathbf{h}_u^{(Z)}\) to the predicted degree \(\hat{d}_u\): 

\begin{equation}
\hat{d}_u = \phi_{str}\left(\mathbf{h}_u^{(Z)}\right).
\label{eq:structure_decoder}
\end{equation} The reconstruction loss for node \(u\) is computed via the L2-distance between the predicted and ground-truth degrees:  

\begin{equation}
\mathcal{L}_u^d = \left\| \hat{d}_u - d_u \right\|_2^2,
\label{eq:node_degree_loss}
\end{equation} where \(d_u = |\mathcal{N}_u|\) is the true degree of node \(u\).

\subsubsection{Neighbor Decoder}  
The deployed neighbor decoder \(\phi_{nbh}(\cdot)\) consists of two collaborative components. To reconstruct the neighbors' feature distribution, we parameterize a multivariate Gaussian \(\mathcal{N}(\hat{\boldsymbol{\mu}}_u, \hat{\boldsymbol{\Sigma}}_u)\) using two MLPs \(\phi_\mu(\cdot)\) and \(\phi_\sigma(\cdot)\), both taking \(\mathbf{h}_u^{(Z)}\) as input:  

\begin{equation}
\hat{\boldsymbol{\mu}}_u = \phi_\mu\left(\mathbf{h}_u^{(Z)}\right), \quad \hat{\boldsymbol{\Sigma}}_u = \text{diag}\left(\exp\left(\phi_\sigma\left(\mathbf{h}_u^{(Z)}\right)\right)\right).
\label{eq:neighborhood_decoder}
\end{equation}
The empirical mean \(\boldsymbol{\mu}_u\) and covariance \(\boldsymbol{\Sigma}_u\) of the neighbors' features are computed as:  
\begin{align}
\boldsymbol{\mu}_u &= \frac{1}{d_u} \sum_{v \in \mathcal{N}_u} \mathbf{h}_v^{(0)} \\ \boldsymbol{\Sigma}_u &= \frac{1}{d_u - 1} \sum_{v \in \mathcal{N}_u} \left(\mathbf{h}_v^{(0)} - \boldsymbol{\mu}_u\right)\left(\mathbf{h}_v^{(0)} - \boldsymbol{\mu}_u\right)^\top + \epsilon\mathbf{I}\,,
\label{eq:neighborhood_representation}
\end{align}
where \(\epsilon > 0\) ensures numerical stability, and \(\mathbf{h}_v^{(0)}\) is the representation of the node \(v\) reside in one-hop neighborhood of node \(u\), i.e. \(v \in \mathcal{N}_u\).
The reconstruction loss follows the KL-divergence formula~\cite{roy2024gad}:  
\begin{multline}
\mathcal{L}_u^n = \frac{1}{2}\left[ \log\frac{|\hat{\boldsymbol{\Sigma}}_u|}{|\boldsymbol{\Sigma}_u|} - p  + \text{tr}\left(\hat{\boldsymbol{\Sigma}}_u^{-1}\boldsymbol{\Sigma}_u\right) \right.\\
\left. + \left(\boldsymbol{\mu}_u - \hat{\boldsymbol{\mu}}_u\right)^\top \hat{\boldsymbol{\Sigma}}_u^{-1}\left(\boldsymbol{\mu}_u - \hat{\boldsymbol{\mu}}_u\right) \right],
\label{eq:neighbor_kl_loss}
\end{multline}
where \(p\) denotes dimension of node representations.

\subsubsection{Attribute Decoder}

We utilize a GDN, which offers greater expressive power and is better suited to fully exploit the inherent structure and properties of graph data~\cite{li2020graph,li2021deconvolutional,cheng2023wiener}. Specifically, we borrow the idea from Wiener Graph Deconvolutional Network~\cite{cheng2023wiener}.

Given an input signal \(\mathbf{x}\) corrupted by additive latent augmentation noise \(\boldsymbol{\epsilon}\) (assumed i.i.d. with \(\mathbb{E}[\epsilon_i] = 0\) and \(\mathbb{VAR}[\epsilon_i] = \sigma^2\)), the Wiener filter minimizes the mean squared error (MSE) in the spectral domain. The reconstruction MSE is formulated as:  
\begin{equation}
\text{MSE}(\hat{\mathbf{x}}) = \sum_{i=1}^N \left( (g_d(\lambda_i)g_c(\lambda_i) - 1 \right)^2 \mathbb{E}[x_i^{*2}] + g_d^2(\lambda_i)\sigma^2,
\label{eq:recon_mse}
\end{equation}
where \(x_i^* = \mathbf{U}^T\mathbf{x}\) represents the spectral projection of input features, \(g_c\) is the graph convolution filter, and \(g_d\) denotes the deconvolution filter to be optimized. By setting the derivative of the MSE with respect to \(g_d(\lambda_i)\) to zero, the optimal deconvolution filter in the spectral domain is derived as:  
\begin{equation}
g_w(\lambda_i) = \frac{g_c(\lambda_i)}{g_c^2(\lambda_i) + \sigma^2 / \mathbb{E}[x_i^{*2}]}\,,
\label{eq:decon_filter}
\end{equation}
where \(\sigma^2 / \mathbb{E}[x_i^{*2}]\) is termed the Augmentation-to-Energy Ratio (AER). This ratio dynamically balances noise suppression (dominant in high-AER spectral components) and signal preservation (emphasized in low-AER components).  To improve computational efficiency and facilitate deployment, we adopt Heat kernel diffusion, defined as \(g_c(\lambda_i) = e^{-\lambda_i}\).

To avoid computationally expensive eigendecomposition (\(\mathbf{U}\boldsymbol{\Lambda}\mathbf{U}^T\)), the Wiener kernel is approximated using Remez polynomials. For a continuous function \(\zeta(\lambda)\) defined on \(\lambda \in [0, 2]\), the \(K^{\text{th}}\)-order Remez polynomial \(p_K(\lambda)\) is constructed to minimize the approximation error:  
\begin{equation}
\mathbf{D}_\gamma = \mathbf{U} \left( \sum_{k=0}^K c_{k,\gamma} \lambda^k \right) \mathbf{U}^T = \sum_{k=0}^K c_{k,\gamma} \mathbf{L}^k,
\label{eq:remez}
\end{equation}
where coefficients \(c_{k,\gamma}\) are optimized via interpolation at Chebyshev nodes. This approach reduces the computational complexity to \(O(K|E|)\), making it scalable for large graphs.  

To enhance robustness against perturbations, Gaussian noise $\mathbf{E}$ is injected into the latent embeddings \(\mathbf{H}^{(Z)}\)$\hat{\mathbf{H}}^{(Z)} = \mathbf{H}^{(Z)} + \beta \mathbf{E}, \quad \mathbf{E} \sim \mathcal{N}(\mathbf{0}, \sigma_P^2 \mathbf{I})$,
where \(\sigma_P^2 = \mathbb{VAR}[\mathbf{H}^{(Z)}]\) is the variance of the latent representations, and \(\beta\) controls the noise magnitude. This augmentation implicitly models joint perturbations in both graph topology and features~\cite{cheng2023wiener}. The decoder employs a multi-layer deconvolution architecture with \(Z\) layers, matching the number of layers in the encoder, and \(Q\) Wiener kernel channels per layer. For layer \(i = 1, \dots, Z\) and channel \(q = 1, \dots, Q\):  
\begin{equation}
\begin{split}
\mathbf{Z}_q^{(i-1)} &= \phi\left( \mathbf{D}_{\gamma_q}^{(i)} \hat{\mathbf{H}}^{(i)} \mathbf{W}_q^{(i)} \right), \\
\hat{\mathbf{H}}^{(i-1)} &= \text{AGG}\left( [\mathbf{Z}_1^{(i-1)}, \dots, \mathbf{Z}_Q^{(i-1)}] \right),
\end{split}
\label{eq:multi-layer}
\end{equation}
where \(\mathbf{D}_{\gamma_q}^{(i)}\) indicates adaptive Wiener kernel at layer \(i\), approximated via Remez polynomials, \(\text{AGG}(\cdot)\) denotes aggregation operator, such as summation, and \(\phi\) is non-linear activation functions. The final reconstructed attributes \(\hat{\mathbf{H}}^{(0)}\) are obtained by progressively reversing the encoder’s operations. The model is trained by minimizing the sum of L2-distance between the original attribute \(\mathbf{h}_u^{(0)}\) and its reconstruction representation \(\hat{\mathbf{h}}_u^{(0)}\) across the node set \(\mathcal{V}\):  
\begin{equation}
\mathcal{L}_{u}^{x} = \|\mathbf{h}_u^{(0)} - \hat{\mathbf{h}}_u^{(0)}\|_2\,,
\label{eq:attribute_loss}
\end{equation}
where \(\hat{\mathbf{h}}_u^{(0)}\) is the reconstructed attribute of the
node \(u\) by Wiener Graph Decoder. This loss function ensures stable training by penalizing deviations in both low- and high-frequency spectral components, guided by the Wiener filter’s theoretical guarantees on MSE minimization~\cite{cheng2023wiener}.  

\subsection{Optimization Objective}

The overall reconstruction loss is a combination of the losses to reconstruct the node degree in Eq. \ref{eq:node_degree_loss}, the neighbors' representation distributions in Eq. \ref{eq:neighbor_kl_loss}, and the node self attributes in Eq. \ref{eq:attribute_loss}:

\begin{equation} \label{eq:Optimization Objective}
    \mathcal{L}=\sum_{u \in \mathcal{V}} \mathcal{L}_{u}^{\prime}, \text { where } \mathcal{L}_{u}^{\prime} = \lambda_{d} \mathcal{L}_{u}^{d}+\lambda_{n} \mathcal{L}_{u}^{n}+\lambda_{x} \mathcal{L}_{u}^{x}\,.
\end{equation}
Here, $\lambda_{x}, \lambda_{d}$, and $\lambda_{n}$ are the hyperparameters that control the weights of different types of reconstruction losses.

\section{Experiments and Result Analysis}\label{sec:exp}

The experiments we conduct aim to answer the following research questions:
\begin{itemize}
    \renewcommand{\labelitemi}{}
    \item \textbf{Question 1 (Q1)}: How does our approach perform against carefully selected baseline models?  (see Section \ref{sec:eva})
    \item \textbf{Question 2 (Q2)}: How do different model components and their specific realization affect the performance of our approach? (see Section~\ref{q2})
    \item \textbf{Question 3 (Q3)}: How do the hyperparameters affect the performance of our model? (see Section~\ref{q3}) 
\end{itemize}


\subsection{Experimental Setup}

\subsubsection{Dataset}
We conduct our experiments on five real-world datasets that contain only organic anomalies: Weibo~\cite{weibo}, Reddit~\cite{reddit1,reddit2}, Disney~\cite{books-enron-disney}, Books~\cite{books-enron-disney}, and Enron~\cite{books-enron-disney}. These datasets differ in size and anomaly ratios. A summary is shown in Table \ref{tab:dataset-summary}. 

To better understand the characteristics of the datasets regarding contextual and structural node anomalies, we calculate the neighborhood similarity $N_{sim}$ and the average degree $\overline{deg}$ for both normal and anomalous nodes. $N_{sim}$ quantifies the average similarity between a node and its neighbors by computing the average mean square error of their node feature vectors. $\overline{deg}$ calculates the mean number of connections within a given set of nodes, representing the average number of connections per node. We summarize these characteristics of each dataset in Table \ref{tab:dataset-statistical-summary_undirect}. 
The detailed formulations of $N_{sim}$ and $\overline{deg}$ can be found in Appendix \ref{appendix:datasets}. $\Delta$ represents the difference between normal and anomaly relative to normal \footnote{$\Delta$ is calculated using the formula: $(x_a-x_n)/x_n$ where $x_a$ is Anomaly and $x_n$ is Normal.}.

\begin{table}[!ht]
\caption{Dataset summary with outlier statistics.}
\label{tab:dataset-summary}
\centering
\begin{tabular}{l@{\hspace{6pt}} c@{\hspace{6pt}} c@{\hspace{6pt}} c@{\hspace{6pt}} c@{\hspace{6pt}} c@{\hspace{6pt}} c@{\hspace{6pt}}}
\toprule
\textbf{Dataset} & \textbf{\#Nodes} & \textbf{\#Edges} & \textbf{\#Feat} & $\overline{deg}$ & \textbf{\#Anomaly} &  \textbf{Ratio (\%)} \\ 
\midrule
\texttt{weibo}   & 8,405            & 407,963          & 400             & 48.5                 & \textbf{868}        & \textbf{10.3}       \\
\texttt{reddit}  & 10,984           & 168,016          & 64              & 15.3                 & \textbf{366}        & \textbf{3.3}        \\
\texttt{disney}  & 124              & 335              & 28              & 2.7                  & \textbf{6}          & \textbf{4.8}        \\
\texttt{books}   & 1,418            & 3,695            & 21              & 2.6                  & \textbf{28}         & \textbf{2.0}        \\
\texttt{enron}   & 13,533           & 176,987          & 18              & 13.1                 & \textbf{5}          & \textbf{0.04}       \\

\bottomrule
\end{tabular}
\end{table}

\begin{table}[!ht]
\caption{Statistical summary of the datasets after the graphs are converted to undirected form. A sign is added to $\Delta$ to indicate the direction of change of the anomaly metrics relative to the normal metrics.}
\centering
\label{tab:dataset-statistical-summary_undirect}
\begin{tabular}{l@{\hspace{8pt}} c@{\hspace{6pt}} c@{\hspace{6pt}} c@{\hspace{6pt}} c@{\hspace{6pt}} c@{\hspace{6pt}} c@{\hspace{6pt}}}
\toprule
 &\multicolumn{3}{c}{\textbf{$N_{sim}$}} & \multicolumn{3}{c}{\textbf{$\overline{deg}$}} \\
 \textbf{Dataset} & \textbf{Normal} & \textbf{Anomaly}& $\Delta(\%) $ & \textbf{Normal} & \textbf{Anomaly} & $\Delta (\%) $ \\ 
\midrule

\texttt{weibo}   & 0.6143  & 1.5181  &  +147.14& 91.71  & 44.72 & -51.24   \\     
\texttt{reddit}  & 0.0011  & 0.0012  &  +3.83  & 15.40  & 12.38 & -19.61   \\
\texttt{disney}  & 452.7107& 433.9868&  -4.14  & 5.53   & 2.83  & -48.80   \\
\texttt{books}   & 0.0946  & 0.0827  &  -12.58 & 5.24   & 3.86  & -26.37   \\
\texttt{enron}   & 0.0418  & 0.0281  &  -32.83 & 26.15  & 38.60 & +47.60   \\

\bottomrule
\end{tabular}
\end{table}

\subsubsection{Training Details}
\label{subsec:training}

The model for each dataset was trained for 200 epochs with a fixed learning rate of 0.005 using the Adam optimizer~\cite{adam}. To have results comparable with those of \cite{he2024ada}, we perform 10 experiments for each dataset using different seeds and use the experiment results to compute the mean and STD. We fix a two-layer wavelet-based graph encoder, each with a hidden dimension of 32. We employed a grid search to find the optimal combination of hyperparameters $\lambda_{d}$, $\lambda_{n}$, $\lambda_{x}$, $K$, $\beta$ and $S$. 
The details of the hyperparameters and the hyperparameter space can be found in Appendix \ref{apd:hyp}.
We use the Area Under the Receiver Operating Characteristic Curve (AUC-ROC)~\cite{davis2006relationship} as the primary evaluation metric. All experiments are performed on a Linux server with a 3.39 GHz AMD EPYC 7742 64-core Processor and 1 NVIDIA A100 GPU with 80GB memory. 

\subsubsection{Baseline Methods}
We choose various graph anomaly detection models from recent research publications as baseline methods for a comprehensive comparison with GRASPED. We include three non-deep learning methods: SCAN \cite{xu2007scan}, Radar \cite{li2017radar} and ANOMALOUS \cite{peng2018anomalous}. SCAN is a structure-based AD model, while ANOMALOUS and Radar are residual reconstruction-based models that consider both structural and attribute features. For deep learning based methods, we chose an adversarial learning-based method GAAN \cite{chen2020generative} as well as a variety of popular graph autoencoder-based methods, such as MLPAE \cite{kipf2016variational}, GCNAE \cite{kipf2016variational}, DOMINANT \cite{ding2019deep}, DONE \cite{bandyopadhyay2020outlier}, AdONE \cite{bandyopadhyay2020outlier}, AnomalyDAE \cite{fan2020anomalydae}, CONAD \cite{xu2022contrastive} and the more recent methods like GAD-NR \cite{roy2024gad} as well as ADA-GAD \cite{he2024ada}. 

\subsection{Experiment 1: Performance comparison (Q1)}
\label{sec:eva}
In Table\ref{tab:overall-performance}, we show the results of GRASPED on the benchmark anomaly detection with baseline models, including prior SOTA models. We reported the baseline results following \cite{he2024ada}, presenting the average performance with its standard deviation (STD). Since the results reported by GAD-NR are based on only five experimental runs, we reran the experiments using our setting in Section \ref{subsec:training} with their code\footnote{\url{https://github.com/Graph-COM/GAD-NR}} and the hyperparameters provided in their paper \cite{roy2024gad} to obtain results. From the results, we can observe that GRASPED outperforms the baseline models on most datasets in detecting benchmark anomaly labels. 

In Table \ref{tab:opt_hyp} we present the optimal parameters for each dataset identified using grid search. We observe that, except for the Books dataset, the contribution of the degree loss to the performance is minimal. A possible explanation for this is that the structural information was already incorporated during the computation of the Laplacian.

\begin{table}[!ht]
\caption{Performance (ROC-AUC) (mean ± STD) comparison across datasets for various methods. The first 3 methods are non-deep learning models. The best result is highlighted in bold, while the second best is underlined.}
\label{tab:overall-performance}
\centering

\begin{tabular}{l@{\hspace{2pt}} l@{\hspace{4pt}} l@{\hspace{4pt}} l@{\hspace{4pt}} l@{\hspace{4pt}} l}
\toprule
\textbf{Algorithm} & \textbf{Weibo} & \textbf{Reddit} & \textbf{Disney} & \textbf{Books} & \textbf{Enron} \\ 
\midrule
SCAN                      & 70.6$\pm$0.0  & 49.6$\pm$0.0  & 50.8$\pm$0.0  & 52.4$\pm$0.0  & 53.7$\pm$0.0  \\ 
Radar                     & \underline{98.2$\pm$0.0}  & 56.6$\pm$0.0  & 50.1$\pm$0.0  & 56.2$\pm$0.0 & 64.1$\pm$0.0  \\ 
ANOMALOUS                & \underline{98.2$\pm$0.0}  & 51.5$\pm$8.6  & 50.1$\pm$0.0  & 52.5$\pm$0.0  & 63.6$\pm$0.3  \\
\midrule
MLPAE                 & 90.0$\pm$0.4  & 49.7$\pm$1.7  & 48.0$\pm$0.0  & 51.2$\pm$5.3  & 41.5$\pm$2.5  \\ 
GCNAE                    & 88.9$\pm$0.3  & 50.7$\pm$0.4  & 47.3$\pm$1.3  & 54.8$\pm$1.5  & 66.8$\pm$0.5  \\ 
DOMINANT                  & 92.1$\pm$0.4 & 56.2$\pm$0.0  & 52.9$\pm$3.0  & 40.1$\pm$2.6  & 54.9$\pm$0.6  \\ 
DONE                    & 86.8$\pm$0.3  & 51.4$\pm$2.2  & 48.4$\pm$4.2  & 54.0$\pm$1.6  & 61.0$\pm$3.1  \\ 
AdONE                   & 82.9$\pm$0.6  & 51.5$\pm$1.3  & 50.9$\pm$2.3  & 54.1$\pm$1.6  & 58.3$\pm$7.3  \\ 
AnomalyDAE                & 92.9$\pm$0.4  & 52.2$\pm$2.0  & 48.2$\pm$4.1  & 59.8$\pm$4.8  & 45.8$\pm$13.1 \\ 
GAAN                     & 92.5$\pm$0.0  & 51.2$\pm$1.1  & 48.0$\pm$0.0  & 53.3$\pm$2.1  & 56.5$\pm$11.6  \\ 
CONAD                    & 90.8$\pm$0.5 & 56.0$\pm$0.0  & 45.3$\pm$4.6  & 40.8$\pm$1.1  & 54.6$\pm$0.4  \\ 
GAD-NR & 78.6$\pm$2.8 &  \underline{58.9$\pm$2.5} & 63.8$\pm$6.7 & \textbf{76.5$\pm$1.7} & \underline{75.3$\pm$4.3} \\ 
ADA-GAD & \textbf{98.4$\pm$0.3} & 56.8$\pm$0.0 & \underline{70.0$\pm$3.0} & 65.2$\pm$3.1 & 72.8$\pm$0.8 \\ 
\midrule
GRASPED & 82.1$\pm$1.3 & \textbf{59.9$\pm$0.2} & \textbf{83.0$\pm$3.6} & \underline{69.1$\pm$2.6} & \textbf{83.3$\pm$3.9} \\ 
\bottomrule
\end{tabular}
\end{table}

\begin{table}[!ht]
\centering
\caption{The optimal hyperparameter for each dataset.}
\label{tab:opt_hyp}
\begin{tabular}{lcccccc}
\toprule
Dataset & $K$  & $\beta$ & $S$ & $\lambda_n$ & $\lambda_x$ & $\lambda_d$ \\ 
\midrule

\texttt{weibo}   & 8  & 0.5 & 45 & 0 & 4 & 0\\          
\texttt{reddit}  & 16  & 0.5 & 35 & 0.2 & 6 & 0\\
\texttt{disney}  & 8  & 1.2 & 20 & 0.6 & 3 & 0\\
\texttt{books}   & 128  & 1.5 & 20 & 3 & 6 & 0.05\\
\texttt{enron}   & 16  & 0.5 & 20 & 0.4 & 3 & 0\\

\bottomrule
\end{tabular}
\end{table}

\subsection{Experiment 2: Ablation Study (Q2)}
To understand and evaluate the contributions of different components of the GAE to GRASPED's performance, we conduct a series of experiments to compare the performance of different combinations of the encoder and feature decoder, while keeping the structure decoder constant. In this study, we explore three different variants of GRASPED, as shown in Table \ref{tab:variants}. 

\label{q2}
\begin{table}[!ht]
\centering
\caption{Different encoder-decoder combinations for the ablation study. Decoder A is the attribute decoder, Decoder S is the structure decoder, and Decoder N is the neighbor decoder.}
\label{tab:variants}
\begin{tabular}{c c@{\hspace{4pt}} c@{\hspace{4pt}} c@{\hspace{4pt}} c}
\toprule
Variant & Encoder & Decoder A& Decoder S& Decoder N\\
\midrule
 1    & Wavelet-based    & MLP-based    & MLP-based    & MLP-based    \\
 2    & GAT-based    & GDN-based    & MLP-based    & MLP-based    \\ 
 3    & GCN-based    & GDN-based   & MLP-based    & MLP-based    \\
\midrule
Ours  & Wavelet-based    & GDN-based   & MLP-based    & MLP-based \\
\bottomrule
\end{tabular}
\end{table}
The results of the experiments are presented in Table \ref{tab:ablation}, with GRASPED's performance appended to the last row. From the results, we observe the following:
\begin{itemize}
    \item GRASPED consistently outperforms all variants across most datasets, while being lower than the best-performing variant on the Weibo dataset. This confirms the effectiveness of using Graph Encoder and GDN as the attribute decoder regardless of the nature of the datasets.
    \item All three variants exhibit degraded performance in different cases, whereas GRASPED achieves the best results on 4 out of the 5 datasets. This indicates the vital role of both the Graph Encoder and the GDN decoder in effectively capturing multi-resolution spectral information and reconstructing graph features consistently over different datasets.
    \item Despite the reduced performance in all three variants, their results outperform the baseline SOTA models from Table \ref{tab:overall-performance} in most datasets. This demonstrates the robustness and general effectiveness of GRASPED’s architecture, even in less optimized configurations.
    \item Table \ref{tab:dataset-statistical-summary_undirect} shows that the differences between anomaly and normal nodes in the Weibo dataset are more pronounced compared to the other datasets. This distinction suggests that employing relatively complex models may cause the overlearning of underlying patterns, leading to suboptimal performance. As shown in Table \ref{tab:ablation}, the performance of Variants 1, 2, and 3 consistently surpasses that of GRASPED on the Weibo dataset. This observation is further supported by the results in Table \ref{tab:overall-performance}, where most non-deep learning models, which consider both contextual and structural information, perform better than most deep learning models - including specialized graph-based deep learning models - on the Weibo dataset. In contrast, more complex models perform better on complex datasets, such as the Reddit dataset, where the contextual and structural differences between normal and anomaly nodes are relatively small.
\end{itemize}
These findings validate the design choices of GRASPED and demonstrate the effectiveness of its Graph Encoder and GDN-based attribute decoder in capturing and utilizing graph structural and spectral information. The ablation study further reinforces the rationality and effectiveness of GRASPED’s architecture.

\begin{table}[!ht]
\centering
\caption{Performance (ROC-AUC) (mean ± STD)  comparison of GRASPED variants in the ablation study. The best result is highlighted in bold, while the second best is underlined.}
\label{tab:ablation}
\begin{tabular}{c@{\hspace{5pt}} c@{\hspace{5pt}} c@{\hspace{5pt}} c@{\hspace{5pt}} c@{\hspace{5pt}} c}
\toprule
\textbf{Variant} & \textbf{Weibo} & \textbf{Reddit} & \textbf{Disney} & \textbf{Books} & \textbf{Enron} \\ 
\midrule

1 & 86.9$\pm$0.8 & 54.5$\pm$1.7 & 80.7$\pm$2.3 & 68.8$\pm$1.1 & 81.4$\pm$3.6 \\
2 & \textbf{90.4$\pm$0.2} & \underline{59.9$\pm$0.5} & \underline{82.7$\pm$4.2} & \underline{65.4$\pm$4.0} & 78.8$\pm$3.5 \\ 
3 & \underline{87.7$\pm$0.83} & 59.0$\pm$0.03 & 80.5$\pm$2.7 & 66.1$\pm$2.6 & \underline{81.6$\pm$4.8} \\
\midrule
Ours & 82.1$\pm$1.3 & \textbf{59.9$\pm$0.2} & \textbf{83.0$\pm$3.6} & \textbf{69.1$\pm$2.6} & \textbf{83.3$\pm$3.9} \\ 
\bottomrule
\end{tabular}
\end{table}

\subsection{Experiment 3: Hyperparameter Analysis (Q3)}
\label{q3}
To analyze the sensitivity of GRASPED to its hyperparameters, we conduct experiments on three key parameters: the maximum dilation parameter \(K\), the augmentation magnitude \(\beta\), and the sample size $S$. For each experiment, we vary the parameter under analysis while keeping the other hyperparameters fixed. Each experiment is conducted 10 times using different seeds to account for the variability of the results. We present the mean results along with the corresponding STD in Figures \ref{fig:hypparam_K}, \ref{fig:hypparam_beta}, and \ref{fig:hyperparam_S}. The standard deviations are omitted from the plots for better visualization. Due to time constraints, the results on the Weibo dataset are excluded from this analysis. Nevertheless, analyzing the remaining four datasets is particularly interesting, as our method performs better on these datasets.

\subsubsection{Analysis on \(K\)}
\begin{figure}[h]
    \centering
    \includegraphics[width=6cm]{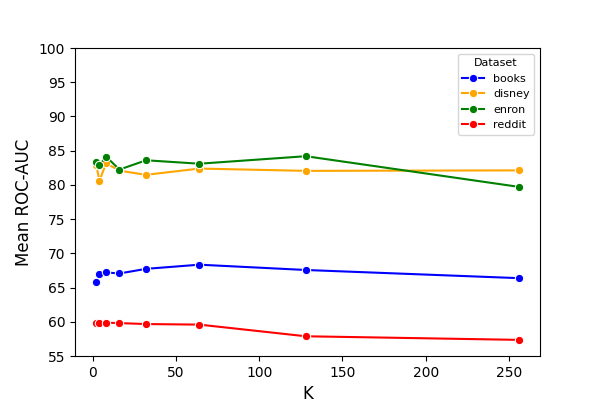}
    \caption{Experiment results of the hyperparameter $K$ for all datasets.}
    \label{fig:hypparam_K}
\end{figure}

The maximum dilation parameter \(K\) is defined as \(K = 2^J\), where \(J\) is the total number of decomposition levels. 
At the \(j\)-th decomposition level, the dilation parameter is \(k = 2^j\). \(K\) determines the number of spectral domain levels explored during the decomposition. The results are presented in Figure \ref{fig:hypparam_K}, where the y-axis represents the ROC-AUC and the x-axis represents \(\log_2 K\). Each data point indicates the mean ROC-AUC, while the shaded region indicates the STD. 

The results suggest that the best performance is generally achieved for intermediate values of \(K\).  
For small \(K\), the encoder fails to adequately decompose spectral domain information, resulting in limited expressiveness. In this case, the graph convolution becomes comparable to a standard GCN, which affects anomaly detection performance. 
For large \(K\), the recursive nature of the wavelet transform excessively splits the low-frequency components, offering diminishing benefits for capturing high-frequency anomalies. This can lead to overfitting, causing ROC-AUC to plateau or even decrease. This phenomenon is particularly evident in the experiments on the Reddit dataset, where $\log_2(K)>6$ leads to a significant drop in performance.

\subsubsection{Analysis on \(\beta\)}
\begin{figure}[h]
    \centering
    \includegraphics[width=6cm]{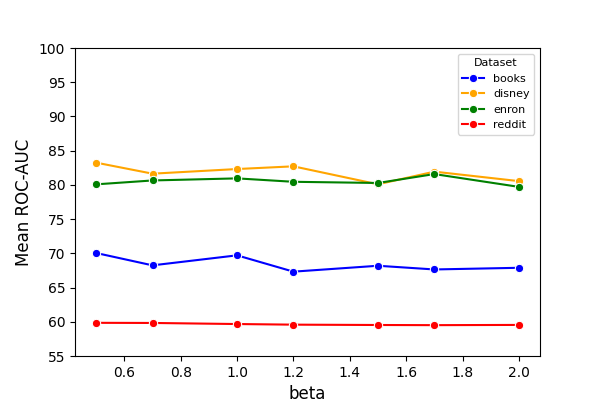}
    \caption{Experiment results of the hyperparameter $\beta$ for all datasets.}
    \label{fig:hypparam_beta}
\end{figure}

The parameter \(\beta\) controls the magnitude of the augmentations applied during training, which influences the robustness of GRASPED. The results of the experiments are shown in Figure \ref{fig:hypparam_beta}. The results indicate that varying \(\beta\) does not significantly affect GRASPED’s ability to detect anomalous nodes. This robustness highlights the model’s reliability and effectiveness, as it can maintain consistent performance irrespective of the specific augmentation magnitude.

\subsubsection{Analysis on $S$}
\begin{figure}[h]
    \centering
    \includegraphics[width=6cm]{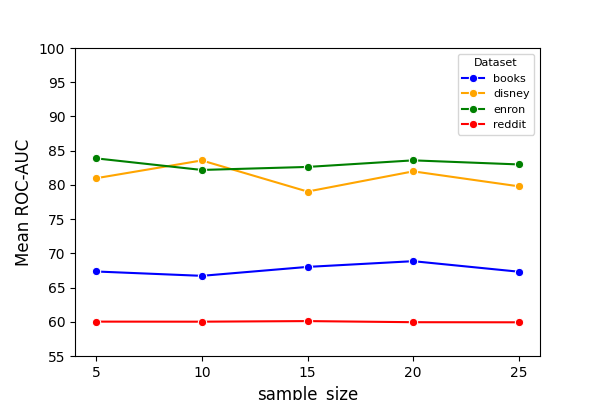}
    \caption{Experiment results of the hyperparameter $S$ for all datasets.}
    \vspace{10pt}
    \label{fig:hyperparam_S}
\end{figure}

The sample size $S$ refers to the number of nodes sampled within a neighborhood to compute distributions. In our experiments, we vary the sample size across \([5, 10, 15, 20, 25]\). The results are summarized in Figure \ref{fig:hyperparam_S}. The findings show no clear trend in performance across four different datasets. However, a larger $S$ can increase computational overhead, especially for larger datasets, suggesting a trade-off between computational cost and model performance. This emphasizes the need for practical experimentation to balance efficiency and effectiveness in real-world applications.

\subsection{Analysis on Cora with Synthetic Anomalies}\label{sec12}
\begin{figure}[h]
    \centering
    \includegraphics[width=6cm]{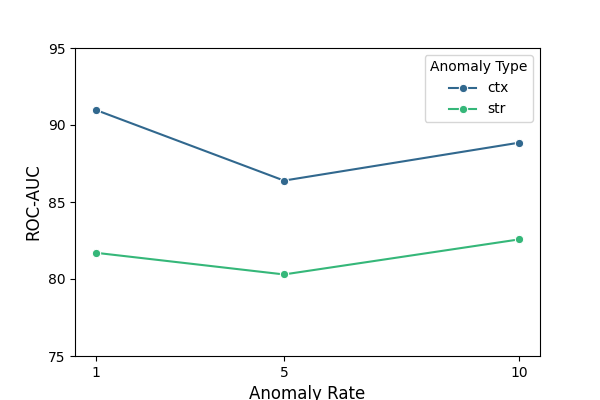}
    \caption{Results on the Cora dataset with different anomaly types and injection rates.}
    \vspace{10pt}
    \label{fig:cora_syn_results}
\end{figure}

To assess our model's performance on different types of anomalies, we perform additional experiments using the Cora dataset, which is injected with two types of anomalies: contextual anomalies (ctx) and structural anomalies (str) as defined in \cite{liu2022bond}. The anomalies are injected into the dataset at different anomaly rates (1\%, 5\% and 10\%) to evaluate how varying anomaly frequencies affect the performance of our model. The results are illustrated in Figure \ref{fig:cora_syn_results}, while the optimal parameters obtained by grid search for each experiment are shown in Figure \ref{tab:opt_hyp_cora}.

In Figure \ref{fig:cora_syn_results}, we observe that the anomaly rate does not significantly impact the overall performance of our model. The ROC-AUC remains stable from 1\% to 10\%, with a slight decrease observed at 5\% for both types of anomalies. This indicates the robustness of our model against different anomaly rates. Furthermore, when comparing the detection capabilities for contextual and structural anomalies, our model performs better in detecting contextual anomalies, which highlights the ability of the model to capture contextual information in graphs.

\begin{table}[!ht]
\centering
\caption{The optimal hyperparameter for Cora with different anomaly types and injection rates.}
\label{tab:opt_hyp_cora}
\begin{tabular}{cccccccc}
\toprule
\textbf{Type} & \textbf{Rate (\%)} & $K$  & $\beta$ & $S$ & $\lambda_n$ & $\lambda_x$ & $\lambda_d$ \\ 
\midrule
ctx & 1 & 16  & 1.5 & 70 & 0.2 & 10 & 0 \\ 
ctx & 5 & 16  & 0.7 & 70 &  0.2 & 10 & 0 \\ 
ctx & 10 & 16  & 0.7 & 70 &  0.2 & 10 & 0 \\ 
str & 1 & 128 & 0 & 35 & 0.6 & 10 & 0 \\ 
str & 5 & 128  & 1.0 & 35 & 0.6 & 3 & 0 \\ 
str & 10 & 128 & 1.5 & 45 & 0.4 & 4 & 0 \\ 

\bottomrule
\end{tabular}
\end{table}

\section{Conclusion}\label{sec13}
In this paper, we introduce GRASPED, a GAE-based unsupervised anomaly detection model. GRASPED utilizes an encoder to capture spectral information at multiple resolutions and a graph deconvolution-based decoder to effectively reconstruct graph embeddings. Experimental results on real-world datasets for node anomaly detection demonstrate that our model achieved state-of-the-art performance on several benchmarks. Further experiments highlight the effectiveness of our model in capturing and utilizing graph structural and spectral information, as well as its stability in performance across different hyperparameters. For future work, we plan to extend GRASPED for direct application on directed graphs, eliminating the need for prior conversion of the graphs to an undirected form. 

\begin{ack}
This work was funded by the Bavarian Ministry of Economic Affairs, Regional Development and Energy in the project “Cognitive Security”.
\end{ack}



\bibliography{mybibfile}

\begin{thebibliography}{43}
\providecommand{\natexlab}[1]{#1}
\providecommand{\url}[1]{\texttt{#1}}
\expandafter\ifx\csname urlstyle\endcsname\relax
  \providecommand{\doi}[1]{doi: #1}\else
  \providecommand{\doi}{doi: \begingroup \urlstyle{rm}\Url}\fi

\bibitem[Bandyopadhyay et~al.(2020)Bandyopadhyay, N, Vivek, and Murty]{bandyopadhyay2020outlier}
S.~Bandyopadhyay, L.~N, S.~V. Vivek, and M.~N. Murty.
\newblock Outlier resistant unsupervised deep architectures for attributed network embedding.
\newblock In \emph{Proceedings of the 13th international conference on web search and data mining}, pages 25--33, 2020.

\bibitem[Chen et~al.(2018)Chen, Yeo, Lee, and Lau]{chen2018autoencoder}
Z.~Chen, C.~K. Yeo, B.~S. Lee, and C.~T. Lau.
\newblock Autoencoder-based network anomaly detection.
\newblock In \emph{2018 Wireless telecommunications symposium (WTS)}, pages 1--5. IEEE, 2018.

\bibitem[Chen et~al.(2020)Chen, Liu, Wang, Dai, Lv, and Bo]{chen2020generative}
Z.~Chen, B.~Liu, M.~Wang, P.~Dai, J.~Lv, and L.~Bo.
\newblock Generative adversarial attributed network anomaly detection.
\newblock In \emph{Proceedings of the 29th ACM International Conference on Information \& Knowledge Management}, pages 1989--1992, 2020.

\bibitem[Cheng et~al.(2023)Cheng, Li, Li, and Tsung]{cheng2023wiener}
J.~Cheng, M.~Li, J.~Li, and F.~Tsung.
\newblock Wiener graph deconvolutional network improves graph self-supervised learning.
\newblock In \emph{Proceedings of the AAAI conference on artificial intelligence}, volume~37, pages 7131--7139, 2023.

\bibitem[Davis and Goadrich(2006)]{davis2006relationship}
J.~Davis and M.~Goadrich.
\newblock The relationship between precision-recall and roc curves.
\newblock In \emph{Proceedings of the 23rd international conference on Machine learning}, pages 233--240, 2006.

\bibitem[Defferrard et~al.(2016)Defferrard, Bresson, and Vandergheynst]{defferrard2016convolutional}
M.~Defferrard, X.~Bresson, and P.~Vandergheynst.
\newblock Convolutional neural networks on graphs with fast localized spectral filtering.
\newblock \emph{Advances in neural information processing systems}, 29, 2016.

\bibitem[Ding et~al.(2019)Ding, Li, Bhanushali, and Liu]{ding2019deep}
K.~Ding, J.~Li, R.~Bhanushali, and H.~Liu.
\newblock Deep anomaly detection on attributed networks.
\newblock In \emph{Proceedings of the 2019 SIAM international conference on data mining}, pages 594--602. SIAM, 2019.

\bibitem[Fan et~al.(2020)Fan, Zhang, and Li]{fan2020anomalydae}
H.~Fan, F.~Zhang, and Z.~Li.
\newblock Anomalydae: Dual autoencoder for anomaly detection on attributed networks.
\newblock In \emph{ICASSP 2020-2020 IEEE International Conference on Acoustics, Speech and Signal Processing (ICASSP)}, pages 5685--5689. IEEE, 2020.

\bibitem[Hammond et~al.(2011)Hammond, Vandergheynst, and Gribonval]{hammond2011wavelets}
D.~K. Hammond, P.~Vandergheynst, and R.~Gribonval.
\newblock Wavelets on graphs via spectral graph theory.
\newblock \emph{Applied and Computational Harmonic Analysis}, 30\penalty0 (2):\penalty0 129--150, 2011.

\bibitem[He et~al.(2024)He, Xu, Jiang, Wang, and Huang]{he2024ada}
J.~He, Q.~Xu, Y.~Jiang, Z.~Wang, and Q.~Huang.
\newblock Ada-gad: Anomaly-denoised autoencoders for graph anomaly detection.
\newblock In \emph{Proceedings of the AAAI Conference on Artificial Intelligence}, volume~38, pages 8481--8489, 2024.

\bibitem[Heard et~al.(2010)Heard, Weston, Platanioti, and Hand]{Heard_2010}
N.~A. Heard, D.~J. Weston, K.~Platanioti, and D.~J. Hand.
\newblock Bayesian anomaly detection methods for social networks.
\newblock \emph{The Annals of Applied Statistics}, 4\penalty0 (2), June 2010.
\newblock ISSN 1932-6157.
\newblock \doi{10.1214/10-aoas329}.
\newblock URL \url{http://dx.doi.org/10.1214/10-AOAS329}.

\bibitem[Kingma and Ba(2014)]{adam}
D.~P. Kingma and J.~Ba.
\newblock Adam: A method for stochastic optimization.
\newblock \emph{arXiv preprint arXiv:1412.6980}, 2014.

\bibitem[Kipf and Welling(2016{\natexlab{a}})]{kipf2016semi}
T.~N. Kipf and M.~Welling.
\newblock Semi-supervised classification with graph convolutional networks.
\newblock \emph{arXiv preprint arXiv:1609.02907}, 2016{\natexlab{a}}.

\bibitem[Kipf and Welling(2016{\natexlab{b}})]{kipf2016variational}
T.~N. Kipf and M.~Welling.
\newblock Variational graph auto-encoders.
\newblock \emph{arXiv preprint arXiv:1611.07308}, 2016{\natexlab{b}}.

\bibitem[Kumar et~al.(2019)Kumar, Zhang, and Leskovec]{reddit1}
S.~Kumar, X.~Zhang, and J.~Leskovec.
\newblock Predicting dynamic embedding trajectory in temporal interaction networks.
\newblock In \emph{Proceedings of the 25th ACM SIGKDD international conference on knowledge discovery \& data mining}, pages 1269--1278, 2019.

\bibitem[Li et~al.(2017)Li, Dani, Hu, and Liu]{li2017radar}
J.~Li, H.~Dani, X.~Hu, and H.~Liu.
\newblock Radar: Residual analysis for anomaly detection in attributed networks.
\newblock In \emph{IJCAI}, volume~17, pages 2152--2158, 2017.

\bibitem[Li et~al.(2020)Li, Yu, Juan, Gopalan, Cheng, and Tomkins]{li2020graph}
J.~Li, T.~Yu, D.-C. Juan, A.~Gopalan, H.~Cheng, and A.~Tomkins.
\newblock Graph autoencoders with deconvolutional networks.
\newblock \emph{arXiv preprint arXiv:2012.11898}, 2020.

\bibitem[Li et~al.(2021)Li, Li, Liu, Yu, Li, and Cheng]{li2021deconvolutional}
J.~Li, J.~Li, Y.~Liu, J.~Yu, Y.~Li, and H.~Cheng.
\newblock Deconvolutional networks on graph data.
\newblock \emph{Advances in Neural Information Processing Systems}, 34:\penalty0 21019--21030, 2021.

\bibitem[Liu et~al.(2021{\natexlab{a}})Liu, Sun, Ao, Feng, He, and Yang]{liu2021intention}
C.~Liu, L.~Sun, X.~Ao, J.~Feng, Q.~He, and H.~Yang.
\newblock Intention-aware heterogeneous graph attention networks for fraud transactions detection.
\newblock In \emph{Proceedings of the 27th ACM SIGKDD conference on knowledge discovery \& data mining}, pages 3280--3288, 2021{\natexlab{a}}.

\bibitem[Liu et~al.(2022)Liu, Dou, Zhao, Ding, Hu, Zhang, Ding, Chen, Peng, Shu, et~al.]{liu2022bond}
K.~Liu, Y.~Dou, Y.~Zhao, X.~Ding, X.~Hu, R.~Zhang, K.~Ding, C.~Chen, H.~Peng, K.~Shu, et~al.
\newblock Bond: Benchmarking unsupervised outlier node detection on static attributed graphs.
\newblock \emph{Advances in Neural Information Processing Systems}, 35:\penalty0 27021--27035, 2022.

\bibitem[Liu et~al.(2021{\natexlab{b}})Liu, Ao, Qin, Chi, Feng, Yang, and He]{liu2021pick}
Y.~Liu, X.~Ao, Z.~Qin, J.~Chi, J.~Feng, H.~Yang, and Q.~He.
\newblock Pick and choose: a gnn-based imbalanced learning approach for fraud detection.
\newblock In \emph{Proceedings of the web conference 2021}, pages 3168--3177, 2021{\natexlab{b}}.

\bibitem[Liu et~al.(2020)Liu, Dou, Yu, Deng, and Peng]{liu2020alleviating}
Z.~Liu, Y.~Dou, P.~S. Yu, Y.~Deng, and H.~Peng.
\newblock Alleviating the inconsistency problem of applying graph neural network to fraud detection.
\newblock In \emph{Proceedings of the 43rd international ACM SIGIR conference on research and development in information retrieval}, pages 1569--1572, 2020.

\bibitem[Mallat(1999)]{mallat1999wavelet}
S.~Mallat.
\newblock \emph{A wavelet tour of signal processing}.
\newblock Elsevier, 1999.

\bibitem[Nt and Maehara(2019)]{nt2019revisiting}
H.~Nt and T.~Maehara.
\newblock Revisiting graph neural networks: All we have is low-pass filters.
\newblock \emph{arXiv preprint arXiv:1905.09550}, 2019.

\bibitem[Peng et~al.(2018)Peng, Luo, Li, Liu, Zheng, et~al.]{peng2018anomalous}
Z.~Peng, M.~Luo, J.~Li, H.~Liu, Q.~Zheng, et~al.
\newblock Anomalous: A joint modeling approach for anomaly detection on attributed networks.
\newblock In \emph{IJCAI}, volume~18, pages 3513--3519, 2018.

\bibitem[Roy et~al.(2024)Roy, Shu, Li, Yang, Elshocht, Smeets, and Li]{roy2024gad}
A.~Roy, J.~Shu, J.~Li, C.~Yang, O.~Elshocht, J.~Smeets, and P.~Li.
\newblock Gad-nr: Graph anomaly detection via neighborhood reconstruction.
\newblock In \emph{Proceedings of the 17th ACM International Conference on Web Search and Data Mining}, pages 576--585, 2024.

\bibitem[Sakurada and Yairi(2014)]{sakurada2014anomaly}
M.~Sakurada and T.~Yairi.
\newblock Anomaly detection using autoencoders with nonlinear dimensionality reduction.
\newblock In \emph{Proceedings of the MLSDA 2014 2nd workshop on machine learning for sensory data analysis}, pages 4--11, 2014.

\bibitem[S{\'a}nchez et~al.(2013)S{\'a}nchez, M{\"u}ller, Laforet, Keller, and B{\"o}hm]{books-enron-disney}
P.~I. S{\'a}nchez, E.~M{\"u}ller, F.~Laforet, F.~Keller, and K.~B{\"o}hm.
\newblock Statistical selection of congruent subspaces for mining attributed graphs.
\newblock In \emph{2013 IEEE 13th international conference on data mining}, pages 647--656. IEEE, 2013.

\bibitem[Savage et~al.(2016)Savage, Zhang, Yu, Chou, and Wang]{savage2016anomalydetectiononlinesocial}
D.~Savage, X.~Zhang, X.~Yu, P.~Chou, and Q.~Wang.
\newblock Anomaly detection in online social networks, 2016.
\newblock URL \url{https://arxiv.org/abs/1608.00301}.

\bibitem[Shuman et~al.(2013)Shuman, Narang, Frossard, Ortega, and Vandergheynst]{shuman2013emerging}
D.~I. Shuman, S.~K. Narang, P.~Frossard, A.~Ortega, and P.~Vandergheynst.
\newblock The emerging field of signal processing on graphs: Extending high-dimensional data analysis to networks and other irregular domains.
\newblock \emph{IEEE signal processing magazine}, 30\penalty0 (3):\penalty0 83--98, 2013.

\bibitem[Tang et~al.(2022)Tang, Li, Gao, and Li]{tang2022rethinking}
J.~Tang, J.~Li, Z.~Gao, and J.~Li.
\newblock Rethinking graph neural networks for anomaly detection.
\newblock In \emph{International Conference on Machine Learning}, pages 21076--21089. PMLR, 2022.

\bibitem[Veli{\v{c}}kovi{\'c} et~al.(2017)Veli{\v{c}}kovi{\'c}, Cucurull, Casanova, Romero, Lio, and Bengio]{velivckovic2017graph}
P.~Veli{\v{c}}kovi{\'c}, G.~Cucurull, A.~Casanova, A.~Romero, P.~Lio, and Y.~Bengio.
\newblock Graph attention networks.
\newblock \emph{arXiv preprint arXiv:1710.10903}, 2017.

\bibitem[Wang et~al.(2019)Wang, Lin, Cui, Jia, Wang, Fang, Yu, Zhou, Yang, and Qi]{wang2019semi}
D.~Wang, J.~Lin, P.~Cui, Q.~Jia, Z.~Wang, Y.~Fang, Q.~Yu, J.~Zhou, S.~Yang, and Y.~Qi.
\newblock A semi-supervised graph attentive network for financial fraud detection.
\newblock In \emph{2019 IEEE international conference on data mining (ICDM)}, pages 598--607. IEEE, 2019.

\bibitem[Wang et~al.(2021)Wang, Zhang, Guo, Yin, Li, and Chen]{reddit2}
Y.~Wang, J.~Zhang, S.~Guo, H.~Yin, C.~Li, and H.~Chen.
\newblock Decoupling representation learning and classification for gnn-based anomaly detection.
\newblock In \emph{Proceedings of the 44th international ACM SIGIR conference on research and development in information retrieval}, pages 1239--1248, 2021.

\bibitem[Xu et~al.(2019)Xu, Shen, Cao, Qiu, and Cheng]{xu2019graph}
B.~Xu, H.~Shen, Q.~Cao, Y.~Qiu, and X.~Cheng.
\newblock Graph wavelet neural network.
\newblock \emph{arXiv preprint arXiv:1904.07785}, 2019.

\bibitem[Xu et~al.(2007)Xu, Yuruk, Feng, and Schweiger]{xu2007scan}
X.~Xu, N.~Yuruk, Z.~Feng, and T.~A. Schweiger.
\newblock Scan: a structural clustering algorithm for networks.
\newblock In \emph{Proceedings of the 13th ACM SIGKDD international conference on Knowledge discovery and data mining}, pages 824--833, 2007.

\bibitem[Xu et~al.(2022)Xu, Huang, Zhao, Dong, and Li]{xu2022contrastive}
Z.~Xu, X.~Huang, Y.~Zhao, Y.~Dong, and J.~Li.
\newblock Contrastive attributed network anomaly detection with data augmentation.
\newblock In \emph{Pacific-Asia conference on knowledge discovery and data mining}, pages 444--457. Springer, 2022.

\bibitem[Yang et~al.(2024)Yang, Hu, Ouyang, Liu, Wang, Ma, Wang, Su, and Liu]{yang2024wavenet}
Z.~Yang, Y.~Hu, S.~Ouyang, J.~Liu, S.~Wang, X.~Ma, W.~Wang, H.~Su, and Y.~Liu.
\newblock Wavenet: tackling non-stationary graph signals via graph spectral wavelets.
\newblock In \emph{Proceedings of the AAAI Conference on Artificial Intelligence}, volume~38, pages 9287--9295, 2024.

\bibitem[Zhang et~al.(2021)Zhang, Wu, Yang, Beheshti, Xue, Zhou, and Sheng]{zhang2021fraudre}
G.~Zhang, J.~Wu, J.~Yang, A.~Beheshti, S.~Xue, C.~Zhou, and Q.~Z. Sheng.
\newblock { FRAUDRE: Fraud Detection Dual-Resistant to Graph Inconsistency and Imbalance }.
\newblock In \emph{2021 IEEE International Conference on Data Mining (ICDM)}, pages 867--876, Los Alamitos, CA, USA, Dec. 2021. IEEE Computer Society.
\newblock \doi{10.1109/ICDM51629.2021.00098}.
\newblock URL \url{https://doi.ieeecomputersociety.org/10.1109/ICDM51629.2021.00098}.

\bibitem[Zhang et~al.(2022)Zhang, Li, Huang, Wu, Zhou, Yang, and Gao]{zhang2022efraudcom}
G.~Zhang, Z.~Li, J.~Huang, J.~Wu, C.~Zhou, J.~Yang, and J.~Gao.
\newblock efraudcom: An e-commerce fraud detection system via competitive graph neural networks.
\newblock \emph{ACM Trans. Inf. Syst.}, 40\penalty0 (3), Mar. 2022.
\newblock ISSN 1046-8188.
\newblock \doi{10.1145/3474379}.
\newblock URL \url{https://doi.org/10.1145/3474379}.

\bibitem[Zhao et~al.(2020)Zhao, Deng, Yu, Jiang, Wang, and Jiang]{weibo}
T.~Zhao, C.~Deng, K.~Yu, T.~Jiang, D.~Wang, and M.~Jiang.
\newblock Error-bounded graph anomaly loss for gnns.
\newblock In \emph{Proceedings of the 29th ACM International Conference on Information \& Knowledge Management}, pages 1873--1882, 2020.

\bibitem[Zhong et~al.(2024)Zhong, Lin, Zhang, and Xu]{zhong2024survey}
M.~Zhong, M.~Lin, C.~Zhang, and Z.~Xu.
\newblock A survey on graph neural networks for intrusion detection systems: Methods, trends and challenges.
\newblock \emph{Computers \& Security}, page 103821, 2024.

\bibitem[Zhou and Paffenroth(2017)]{zhou2017anomaly}
C.~Zhou and R.~C. Paffenroth.
\newblock Anomaly detection with robust deep autoencoders.
\newblock In \emph{Proceedings of the 23rd ACM SIGKDD international conference on knowledge discovery and data mining}, pages 665--674, 2017.

\end{thebibliography}

\appendix

\section{Property of datasets}
\label{appendix:datasets}
\paragraph{Neighborhood Similarity}
Neighborhood similarity $N_{sim}$ quantifies how similar a node is to its neighbors on average. For a given node $v$, its neighborhood similarity is defined as:

\begin{equation}
N_{sim}(v) = \frac{1}{|N(v)|} \sum_{u \in N(v)} \text{MAE}(v, u),
\end{equation}where:
\begin{itemize}
    \item $N(v)$ is the set of first-order neighbors of node $v$.
    \item $|N(v)|$ is the number of neighbors of $v$.
    \item $\text{MAE}(v, u)$ represents the similarity between nodes $v$ and $u$, which is calculated using the mean absolute error.
\end{itemize}The mean absolute error between two node feature vectors is calculated by:

\begin{equation}
\text{MAE}(v, u) = \frac{1}{n} \sum_{i=1}^{n} |v_i - u_i|.
\end{equation} Our calculation of $N_{sim}$ is performed without normalizing the node feature vectors. 

\paragraph{Average Degree}
The average degree $\overline{deg}$ measures the mean number of connections in a given set of nodes. For a set of nodes $\mathcal{V}$, the average degree is computed as:

\begin{equation}
\overline{deg}(\mathcal{V}) = \frac{1}{|\mathcal{V}|} \sum_{v \in \mathcal{V}} \text{degree}(v),
\end{equation}where:
\begin{itemize}
    \item $V$ represents the set of nodes (which can be all nodes, normal nodes, or anomalous nodes).
    \item $|\mathcal{V}|$ is the number of nodes in the set.
    \item $\text{degree}(\mathcal{V})$ represents the degree of node $v$, i.e., the number of connections it has.
\end{itemize}


\section{Hyperparameters for Grid Search}
\label{apd:hyp}
Specifically, the parameters were explored over the following ranges:
\begin{itemize}
    \item \(\lambda^n \in \{0.2, 0.4, 0.6, 2.0, 3.0, 8.0, 9.0\}\)
    \item \(\lambda^a \in \{0.3, 0.4, 0.6, 1.0, 3.0, 4.0, 6.0, 10.0\}\)
    \item \(\lambda^d \in \{0, 0.05, 0.15, 0.25\}\)
\end{itemize}
For the sample size \(S\), values of \(\{5, 6, 8, 10, 15, 20, 25\}\) were applied for most datasets, while a broader range of \(\{20, 35, 40, 45, 60, 70, 80\}\) was considered when necessary. The number of neighbors \(K\) was varied over \(\{2, 4, 8, 16, 32, 64, 128, 256\}\) and the parameter \(\beta\) was set within \(\{0.3, 0.5, 0.7, 1.0, 1.5\}\). In constructing our attribute decoder based on the Wiener GDN framework~\cite{cheng2023wiener}, we adhered to the default settings provided by Wiener GDN except for the \(\beta\) parameter, which was fine-tuned during training.

\end{document}